\documentclass{article}
\usepackage{algorithm} 
\usepackage{algpseudocode} 
\usepackage{PRIMEarxiv}
\usepackage{adjustbox} 
\usepackage[utf8]{inputenc} 
\usepackage[T1]{fontenc}    
\usepackage{hyperref}       
\usepackage{url}            
\usepackage{booktabs}       
\usepackage{amsfonts}       
\usepackage{nicefrac}       
\usepackage{microtype}      
\usepackage{lipsum}
\usepackage{fancyhdr}       
\usepackage{graphicx}       
\graphicspath{{media/}}     
\usepackage{algorithm} 
\usepackage{algpseudocode}
\usepackage[section]{placeins}
\usepackage{amsmath}
\usepackage{amssymb}
\usepackage{tabularx}
\usepackage{multirow}
\usepackage{calc}
\usepackage{xcolor}
\usepackage{multirow,diagbox,tabularx,blindtext}
\usepackage{threeparttable}
\usepackage{booktabs}    
\usepackage{tabularx}    
\usepackage{xcolor}      
\usepackage{colortbl}    
\usepackage{tcolorbox}   
\usepackage{tikz}        
\usepackage{array} 
\tcbuselibrary{breakable,skins}
\usepackage{lineno}
\usepackage{float}  
\newcommand{\factorsection}[3]{
  \begin{table}[H]  
    \centering
    \begin{tabular}{p{0.95\textwidth}}
      \toprule
      \textbf{\large #1} \\
      \midrule
      \textbf{Summary:} #2 \\
      \midrule
      \textbf{Key Factors:} \\
      \begin{tabular}{@{}p{5cm}r@{}}
        #3
      \end{tabular} \\
      \bottomrule
    \end{tabular}
  \end{table}
}
\title{CrashSage: A Large Language Model-Centered Framework for Contextual and Interpretable Traffic Crash Analysis

}

\author{
   Hao Zhen and  Jidong J. Yang \\
  Smart Mobility and Infrastructure Lab\\
  College of Engineering\\
 University of Georgia, Athens, GA, USA\\
  \texttt{Hao.Zhen, Jidong.Yang@uga.edu} \\
}

\tcbset{breakable,enhanced}
\begin{document}
\maketitle

\begin{abstract}

Road crashes claim over 1.3 million lives annually worldwide and incur global economic losses exceeding \$1.8 trillion. Such profound societal and financial impacts underscore the urgent need for road safety research that uncovers crash mechanisms and delivers actionable insights. Conventional statistical models and tree ensemble approaches typically rely on structured crash data, overlooking contextual nuances and struggling to capture complex relationships and underlying semantics. Moreover, these approaches tend to incur significant information loss, particularly in narrative elements related to  multi-vehicle interactions, crash progression, and rare event characteristics.
This study presents CrashSage, a novel Large Language Model (LLM)-centered framework designed to advance crash analysis and modeling through four key innovations. 
First, we introduce a tabular-to-text transformation strategy paired with relational data integration schema, enabling the conversion of raw, heterogeneous crash data into enriched, structured textual narratives that retain essential structural and relational context.
Second, we apply context-aware data augmentation using a base LLM model to improve narrative coherence while preserving factual integrity. 
Third, we fine-tune the LLaMA3-8B model for crash severity inference, demonstrating superior performance over baseline approaches, including zero-shot, zero-shot with chain-of-thought prompting, and few-shot learning, with multiple models (GPT-4o, GPT-4o-mini, LLaMA3-70B). 
Finally, we employ a gradient-based explainability technique to elucidate model decisions at both the individual crash level and across broader risk factor dimensions. This interpretability mechanism enhances transparency and enables targeted road safety interventions by providing deeper insights into the most influential factors.

\end{abstract}


\section{Introduction}

Traffic crashes remain a persistent global public health crisis, resulting in over 1.3 million fatalities annually and imposing economic costs estimated at more than \$1.8 trillion. In the United States alone, approximately 42,000 lives are lost each year, despite continuous advancements in vehicle safety features, roadway design improvements, and safety policy implementation \cite{NCSA2023}. This enduring challenge highlights the inherently complex and multifaceted nature of traffic crashes, which result from dynamic interactions among human factors, vehicle characteristics, environmental conditions, and infrastructure elements.

The complexity of traffic safety analysis lies in capturing the intricate relationships between these factors. Human factors (e.g.,  variations in physical and physiological status, attentiveness levels, risk-taking behaviors, and social responsibility) interact with dynamic vehicle movements, changing weather conditions, and diverse roadway conditions and characteristics. This complexity necessitates sophisticated analytical approaches that can effectively model these high-dimensional interactions to develop context-aware, targeted interventions and policies.

Over the decades, traffic safety research has evolved through several methodological paradigms, each offering distinct advantages and facing specific limitations. Traditional statistical and econometric models have long served as the cornerstone of crash severity analysis \cite{pervaz2023econometric, peng2017assessing, golob2003relationships, eluru2007joint, savolainen2011statistical,zhen2024unraveling}. These methods, including random parameters multinomial logit models \cite{behnood2016empirical, cerwick2014comparison}, ordered probability models \cite{eluru2015note, yasmin2015analyzing}, latent class models \cite{yasmin2014latent, eluru2012latent}, and markov-switching models \cite{xiong2014analysis}, offer explicit interpretability through parameter estimates that quantify relationships between crash outcomes and related factors.

However, these traditional models impose rigid functional forms and distributional assumptions, which constrain their ability to capture the nonlinearities, complex high-order interactions, and inherent heterogeneity present in crash data \cite{mannering2020big}. These limitations often lead to inconsistent findings across datasets and geographic regions, an issue compounded by the relatively small sample sizes that are common in transportation safety research. Furthermore, specifying appropriate model structures typically requires considerable domain expertise,  introducing potential subjectivity in the selection of variables and the specification of interactions.

In response to these limitations, researchers have increasingly resorted to machine learning (ML) techniques, such as  random forests and deep neural networks, which have demonstrated improved predictive performance in crash modeling \cite{rahim2021deep}. For interpretability, tree-based ensemble models typically rely on post-hoc methods like SHapley Additive exPlanations (SHAP) \cite{lundberg2017unified}, which attribute importance scores to input features. Lares et al. \cite{FGTT} introduces a Feature Group Tabular Transformer (FGTT) model that enhances interpretability by grouping semantically related features into tokens and leveraging transformer attention heatmaps to reveal key interactions between feature groups, thereby uncovering relationships behind different traffic crash types. While these approaches offer greater flexibility in modeling complex, nonlinear relationships, they still suffer from certain limitations. Firstly, they focus on feature-level attribution without revealing the underlying reasoning process. Secondly, they fall short in capturing contextual and causal relationships beyond statistical correlations. Lastly, they lack a native mechanism for processing unstructured data such as textual crash reports, requiring feature engineering to encode relevant information.  This encoding process can lead to substantial information loss when qualitative factors (e.g.,  the sequential nature of crash events, contextual factors, and narrative elements) are reduced to categorical variables or omitted entirely.

Large Language Models (LLM) have emerged as a transformative paradigm in traffic safety analysis, addressing many of aforementioned  limitations inherent in both traditional statistical methods and conventional machine learning techniques. The evolution of language models, from encoder-only architectures like BERT \cite{devlin2018bert} and RoBERTa \cite{liu2019roberta} to decoder-only variants including the GPT series \cite{radford2018improving} and LLaMA \cite{dubey2024llama}, has demonstrated remarkable generalization capabilities across diverse domains~\cite{shi2023chatgraph,shi2025searchrag,shi2023mkrag}, suggesting their potential for transportation safety applications. The unique advantages of LLMs in crash analysis stem from two key strengths: (1) their ability to process and derive insights from unstructured textual narratives, which often contain rich contextual information that is lost in structured tabular formats; and (2) the extensive world knowledge embedded within their pretrained parameters, which may enable nuanced reasoning about complex circumstances described in textual narratives.

Table \ref{tab:comparison_methods} provides a comparative overview of the different methodological approaches to crash severity analysis, highlighting their respective input data requirements, interpretability characteristics, and key limitations. By leveraging unstructured crash narratives, LLMs can preserve the the richness and completeness of crash events by utilizing extensive contextual information, maintaining sequential flow of actions as the event unfolds, and capturing multi-vehicle interactions that are often fragmented or lost in traditional tabular formats. 

Furthermore, LLMs are capable of generating natural language explanations that mirror human reasoning, offering intuitive and comprehensive insights into the dynamics of crash events. Unlike abstract feature importance scores, these explanations are easily interpretable by transportation officials and safety practitioners, even without specialized technical expertise.

Previous research \cite{zhen2024leveraging} suggests that LLM-based crash severity analysis aligns well with domain knowledge while providing competitive accuracy. Various prompting techniques, including zero-shot, few-shot, and chain-of-thought approaches, have shown promise in improving modeling accuracy, while expert validation confirms that LLM-generated explanations correspond with established traffic safety knowledge. Fan et al. \cite{fan2024learning} fine-tuned LLaMA2 \cite{touvron2023llama} for traffic crashes to predict accident outcomes and demonstrated the overall better performance than machine learning baselines: Random forest, Decision Trees, Adaptive boosting (AdaBoost), Bayesian Network(BN), LogisticRegression (LR), and Categorical boosting (CatBoost). It shows the promising application for traffic safety area. However, the limitation exists in the interpretability with intuitive "what-if" analysis. Their explanation method involves making controlled modifications to input data and observing the changes in the model's predictions, which helps to identify the influence of different features on the output. Despite this intuitive method, their approach has two significant limitations that restrict its practical utility in transportation safety applications. First, while their what-if analysis reveals how output distributions shift under hypothetical conditions, it does not expose the internal reasoning process of the model, which is the advantage of LLMs; the transformation of raw data into narrative form and subsequent classification remains largely a black box. Second, the method offers only post-hoc explanations based on output perturbations, rather than providing a detailed, explanation that could enhance transparency and trust. 

This study introduces CrashSage, a novel framework specifically designed to address critical gaps in crash analysis through four principal contributions. First, we develop a comprehensive \emph{tabular-to-text transformation} method along with relational data integration schema that converts raw, heterogeneous crash data from Washington State datasets into richly detailed textual narratives, thereby preserving crucial structural and relational information commonly lost in conventional tabular formats. Second, we implement \emph{context-aware data augmentation} via LLaMA-8B, enhancing the coherence of these crash narratives while rigorously maintaining factual accuracy. Third, we perform \emph{supervised fine-tuning} of a LLaMA3-8B model \cite{llama3modelcard} tailored for crash severity inference, demonstrably surpassing baseline approaches such as zero-shot, zero-shot with chain-of-thought, and few-shot across multiple model configurations (GPT-4o \cite{openai2024gpt4o}, GPT-4o-mini  \cite{openai2024gpt4omini}, and LLaMA3-70B \cite{llama3modelcard}). Finally, we integrate a \emph{gradient-based explainability} strategy to illuminate model decisions at both the individual crash level and in broader risk factor co-occurrence analyses. This interpretability mechanism not only enhances transparency and trustworthiness in model outputs but also provides actionable insights for targeted interventions in road safety management through a deeper understanding of how diverse factors interact to influence crash outcomes.

The CrashSage framework represents a paradigm shift in traffic safety analysis by transforming sparse crash records into actionable narratives. This approach enables transportation agencies to move beyond retrospective statistical analysis toward proactive risk identification and mitigation. By leveraging the semantic understanding capabilities of LLMs while maintaining explainability, our framework bridges the gap between advancement in AI and practical deployment in safety-critical transportation applications.

\begin{table}[h]
    \centering
    \caption{Comparison of methods used in crash severity analysis.}
    \begin{adjustbox}{max width=\textwidth}
    \begin{tabular}{lccc}
        \toprule
        \textbf{Method} & \textbf{Input Data} & \textbf{Interpretability} & \textbf{Limitations} \\
        \midrule
        Econometric and Statistical Methods & Structured data & Explicit, model-based & Assumes fixed functional forms, limited complexity, lacks context \\
        Tree Ensemble Models & Structured data & SHAP-based feature importance & Lacks context, needs feature engineering \\
        Large Language Models (LLMs) & Unstructured data with context & Natural language explanations & Computationally expensive, potential biases\\
        \bottomrule
    \end{tabular}
    \end{adjustbox}
    
    \label{tab:comparison_methods}
\end{table}

\section{Related Work}
This section explores two pivotal domains that underpin our CrashSage framework. First, we examine the evolution of Large Language Models (LLMs) and their emerging applications in transportation safety, emphasizing both technical advancements and domain-specific implementations. Second, we investigate interpretability techniques for language models, with a particular focus on gradient-based attribution methods that offer meaningful explanations in safety-critical contexts. Together, these complementary  areas of research establish the technical foundations of our framework and highlight the key gaps it addresses in advancing explainable LLM capabilities for traffic safety analysis.

\subsection{LLMs and Their Applications in Transportation Safety}

Since their inception, LLMs have rapidly progressed from niche research tools to versatile, general-purpose systems with wide-ranging applications across diverse domains. The development of LLMs is driven by transformer-based architectures. BERT \cite{devlin2018bert} employed encoder-only design well-suited for tasks like classification and information extraction. This was followed by a shift toward decoder-only architectures, exemplified by models such as GPT \cite{radford2018improving},  LLaMA-3 \cite{dubey2024llama}, Claude \cite{anthropic2023claude}, and GPT-4 \cite{openai2023gpt4}.  This evolution has been characterized by notable improvements in contextual reasoning, factual accuracy, and adaptability, positioning LLMs as versatile tools for complex technical applications. The decoder-only models are particularly effective at generating coherent narratives, modeling sequential events, and capturing long-range dependencies. These capabilities are especially valuable for analyzing traffic crash sequences, which could unfold as temporally linked chains of events. Additionally, decoder-based LLMs exhibit strong few-shot and zero-shot learning capabilities, enabling effective generalization to new crash scenarios even in the absence of domain-specific training examples \cite{zhen2024leveraging}. Their ability to handle increasingly longer contexts enables the integration of comprehensive information about road conditions, weather, vehicle states, and driver behaviors within a unified analysis framework. 

The transportation sector has begun to explore the potential of LLMs across a range of  applications, though their adoption in safety-critical contexts remains limited. Recent studies have demonstrated promising use cases, such as traffic forecasting \cite{ren2024tpllm}. Expanding on these advancements, Jonnala et al. \cite{jonnala2025exploring} investigated the role of LLMs in optimizing transit operations, focusing on enhanced route planning, reduced wait times, and personalized travel assistance. By leveraging GTFS data alongside advanced natural language processing techniques, the research demonstrates that, with careful engineering and fine-tuning, LLMs can significantly improve resource allocation and passenger satisfaction. These findings highlight the potential of LLMs to support data-driven decision-making in urban transit systems.

In the specific domain of transportation safety, early applications have focused on analyzing accident reports and extracting structured information from unstructured narratives. Notably, Zhen et al. \cite{zhen2024leveraging} demonstrated that LLMs can effectively classify crash severity from textual descriptions while providing explanations that align with domain knowledge. This approach marks a stark departure from traditional methods that rely exclusively on structured tabular data. 

Despite these promising developments, existing applications of LLMs in transportation safety face several limitations. First, most approaches rely on general-purpose models without domain-specific fine-tuning, limiting their ability to accurately interpret specialized terminology and contextual nuances unique to transportation domain. Second, LLMs are often employed as isolated analytical tools rather than being embedded within integrating frameworks that span the entire decision-making pipeline, from data preparation and feature engineering to actionable insights and policy support. Most notably, current efforts rarely address the explainability requirements that are essential for safety-critical contexts. In such settings, understanding the rationale behind a model's output  is important, particularly when decisions may directly impact public safety.

These limitations underscore the need for transportation-specific adaptations of LLMs that incorporate domain knowledge, systematic data transformation, and robust explainability mechanisms. Addressing these challenges is essential for advancing LLMs from experimental use cases to practical tools that support evidence-based, actionable decision-making in safety-critical transportation contexts. 

\subsection{Explanation in Language Models}

Unlike traditional statistical models with explicit parameters that directly link input features to predictions, LLMs rely on distributed representations that pose significant challenges for interpretation~\cite{wu2024usable}. This section reviews the evolution of explainability approaches for LLMs, with a particular focus on gradient-based methods that inform the design of our CrashSage framework.

Early explanation methods for language models primarily relied on attention visualization \cite{clark2019does}, which offered intuitive but often misleading insights into model behavior. While attention maps can highlight tokens a model emphasizes during processing, subsequent research has shown that attention weights alone do not provide a reliable causal account of model predictions \cite{jain2019attention}. This limitation is especially problematic in transportation safety contexts, where accurately identifying causal factors is essential for designing effective interventions and informing policy decisions.

To overcome these shortcomings, more recent approaches have employed influence functions \cite{han2020explaining} and integrated gradients \cite{sundararajan2017axiomatic} to attribute model predictions to specific input tokens. These methods provide more reliable explanations by quantifying the impact of input perturbations on model outputs. In safety-critical domains like crash analysis, these techniques can potentially identify which aspects of an incident description most significantly influence severity predictions, thereby aligning with traditional safety analysis objectives focused on uncovering key risk factors.

Our CrashSage framework builds upon the comprehensive explanation techniques introduced by Wu et al. \cite{wu2024language}, incorporating gradient-based attribution to assess the influence of input tokens on specific model outputs. This approach uses a Taylor approximation to quantify how the inclusion or exclusion of individual input tokens affect output probabilities. By normalizing these attribution scores, the method yields a robust measure of each input token's contribution to the model predictions. When applied to crash analysis, this technique can highlight key terms within crash descriptions (e.g., "high speed," "distracted," "intersection") that strongly influence severity inference. These insights provide transportation safety professionals with a clearer understanding of the factors driving model decisions, supporting more informed and interpretable use of LLMs in the transportation safety domain.

Adapting explanation methods to transportation safety requires domain-specific considerations that go beyond general text analysis. Crash narratives often contain specialized terminology, structured inter-dependencies, and causal sequences that must be interpreted through the lens of traffic safety principles.  To address these nuances, our framework extends the methods proposed by Wu et al.'s \cite{wu2024language} by adjusting key hyperparameters and fine-tuning the model to align with established safety analysis practices. These adaptations ensure that the resulting explanations accurately reflect model behavior while offering actionable insights for transportation safety professionals.

The interpretability approach in our framework operates across multiple levels of granularity. At the individual crash level, token-level attributions identify critical factors influencing severity predictions, such as vehicle types, environmental conditions, and driver characteristics. At a broader scale, analyzing attribution patterns across multiple incidents enables to reveal systemic safety issues, particularly through co-occurrence analysis of high-attribution factors. This multi-level strategy addresses key interpretability gaps in current LLM applications within transportation safety.  As demonstrated in Section \ref{sec:interpretability}, the visualization of co-occurring factors reveals complex inter-dependencies among environmental conditions, driver behaviors, vehicle attributes, and infrastructure features that contribute to crash outcomes. This approach moves beyond isolated factor analysis to uncover compound effects and  interaction dynamics, providing transportation agencies a more holistic understanding of crash causation mechanisms, which is critical for developing targeted and effective safety interventions.

\section{Data Sources and Processing}

This study leverages Washington traffic crash datasets, sourced from the Washington State Department of Transportation (WSDOT), encompassing crash records from 2020 to 2022. It has four different table, including crash table, road segment table, vehicle/unit table, and person table. Transportation crash data is typically stored in structured relational databases with complex schemas that separate information across multiple tables. While this structure was designed for storage and querying, it presents challenges for natural language processing applications. To address this, we utilized relational database schema combined with tabular-to-text transformation technique, introduced in the subsequent section, to convert structured crash data into coherent narrative descriptions.

\subsection{Relational Schema for Crash Data Integration}
\label{Relational Schema}
\begin{figure}[htbp!]
  \centering
  \includegraphics[width=\textwidth, trim=20pt 370pt 20pt 105pt, clip]{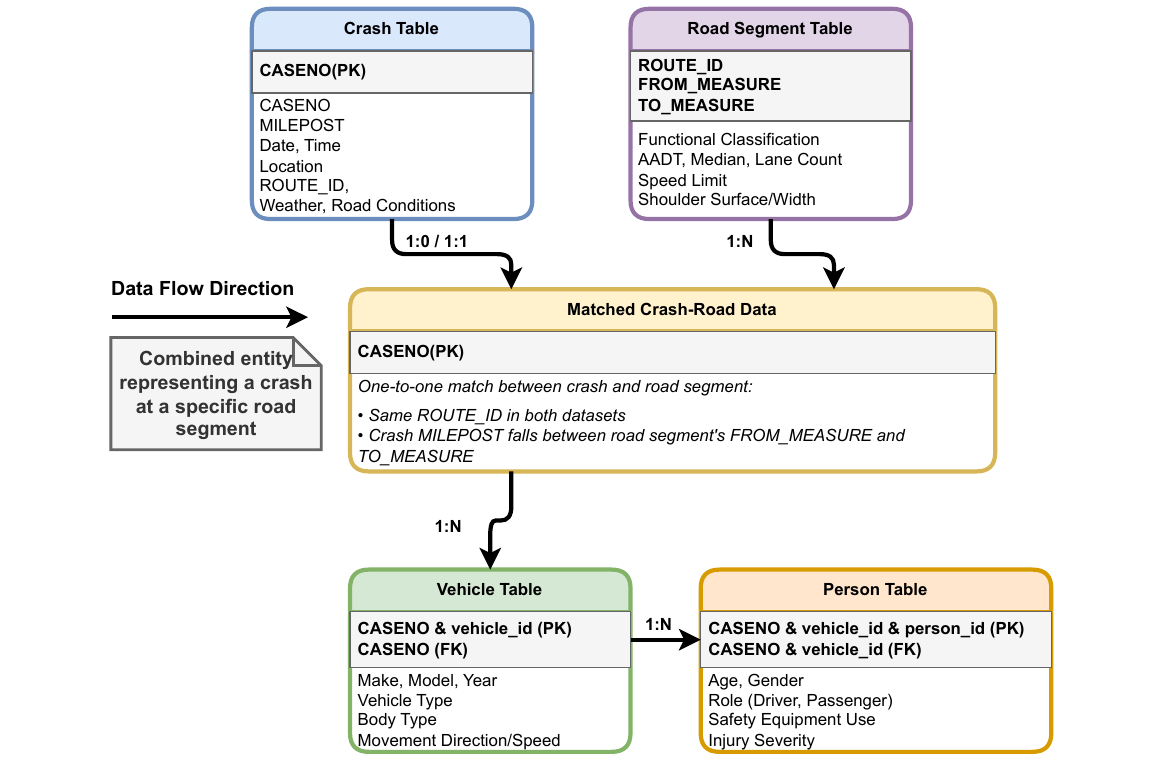}
  \caption{Data integration through relational schema}
  \label{fig:traffic_safety_framework}
\end{figure}

Our framework employs a relational schema, detailed in Figure~\ref{fig:traffic_safety_framework}) to integrate heterogeneous crash data sources through four normalized tables: Crash, Road Segment, Vehicle, and Person. The crash table forms the core entity with primary key CASENO, containing spatiotemporal attributes (location, timestamp) and environmental conditions. Through foreign key relationships, each crash record links to its corresponding Road Segment via spatial matching between MILEPOST and segment boundaries (FROM\_MEASURE, TO\_MEASURE), ensuring accurate geolocation mapping.

The schema maintains data integrity through hierarchical one-to-many relationships: each crash record connects to multiple vehicle entries through CASENO, and each vehicle/unit links to multiple person records. This structure preserves the natural hierarchy of crash events while enabling efficient querying of participant-level details. We implemented this schema through a nested dictionary structure, with JSONL serialization facilitating language model integration through instruction-based learning templates.

Initial analysis revealed a severe disparity between 'No Apparent or Minor Injury' cases (\(n=49,648\)) and 'Serious injury or fatal' incidents (\(n=1,779\)).  To address the class imbalance, we employed a stratified down-sampling approach, resulting in a more balanced dataset, comprising  2,654 no apparent or minor injury cases and 1,779 serious injury or fatal cases. 

\subsection{Tabular-to-Text Transformation}
\label{Tabular-to-Text}
The tabular-to-text conversion pipeline transforms structured crash records into natural language narratives through a two-phase process: semantic normalization and template-based generation. In the normalization phase, numerical codes are mapped to natural language descriptors using domain-specific lexicons, translating categorical encoding into human-interpretable terms while removing duplicates and non-informative null values. The subsequent template-based generation uses fill-in-the-blank templates, illustrated below, to construct coherent, chronological event sequences that highlight key crash dynamics, enabling the generation of context-rich crash narratives suitable for analysis by LLMs.
 
 \begin{quote}
  On [date], a [day of week] at [time], an accident involving [number] vehicles occurred [lighting conditions], with [weather conditions]. The road condition at the time was [surface condition]. The accident took place on [road name] ([road type])...
\end{quote}

Particularly, we structurally separate descriptive narratives (pre-crash conditions, collision mechanics) from outcome related narratives (injury severity, vehicle damage). This division support supervised learning objectives and enables models to learn potential causal relationships between antecedent conditions and resulting consequences.

Consequently, the transformation yields LLM-readable narratives that preserve relational semantics through natural language encoding. This methodology effectively bridges structured crash analytics with unstructured text processing capabilities, facilitating direct application of LLMs to transportation safety analysis while maintaining computational tractability.

\section{CrashSage Framework}
\subsection{System Architecture Overview}

\begin{figure}[ht!]
  \centering
  \includegraphics[width=1\textwidth]{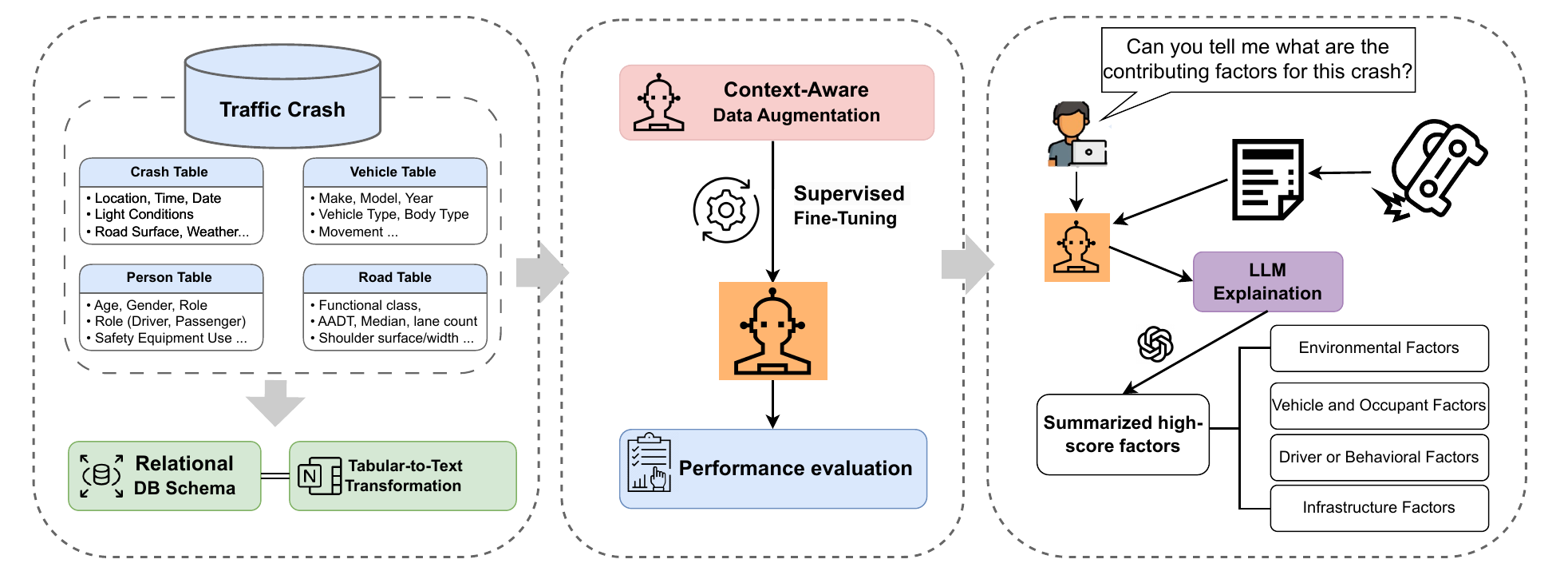}
  \caption{CrashSage framework.}
  \label{fig:safety-assistant}
\end{figure}

Our CrashSage framework presented in Figure~\ref{fig:safety-assistant} offers a comprehensive approach to traffic safety analysis through the integration of LLMs with structured crash data. At its foundation, the framework utilizes entity-oriented analysis, leveraging the \textit {Crash, Vehicle, Person, and Road/Unit} tables, each containing specific attributes that characterize different aspects of crash events. These structured data sources undergo a sequential transformation process beginning with a relational database schema that organizes the hierarchical relationships between entities (Section~\ref{Relational Schema}), followed by a Tabular-to-Text Transformation that converts structured records into coherent narratives (described previously in Section~\ref{Tabular-to-Text}). The resulting template-based narratives are further refined through context-aware data augmentation using a pretrained LLM agent (referred to as the base LLM in this paper), which improves textual coherence while preserving factual accuracy. The augmented narratives serve as the training data for the following supervised fine-tuning of our CrashSage agent, another  LLM specialized in traffic safety domain. Additionally, the analytical capabilities of our framework are enhanced through gradient-based explainability methods, which offer interpretable insights via attribution analysis. This enables the generation of  entity- or aspect-aware explanations, focusing on contributing factors grouped in to key categories: environmental conditions, vehicle and occupant characteristics, driver behavior, and infrastructure related elements, as shown in the right panel of Figure~\ref{fig:safety-assistant}.

The core components of our CrashSage framework, including \textit{Context-Aware Data Augmentation, Supervised Fine-Tuning, and Gradient-Based Explanation}, are detailed in the following sections.

\subsection{Context-Aware Data Augmentation}

Our framework implements a semantic preservation augmentation technique leveraging the LLaMA3-8B model \cite{llama3modelcard} to enhance narrative coherence while maintaining factual fidelity in crash reports. 
The augmentation pipeline operates through a linguistic transformation process formalized as:


\begin{equation}
  R = \text{LLM}(x, \{c_i\}_{i=1}^n), p_t),
\end{equation}

where \( R \) represents the enhanced narrative,
and \( \{c_i\} \) defines a set of preservation constraints. The model receives specialized instructions $p_t$ through system prompts that position it as a domain expert in transportation safety reporting. We employ a conservative temperature setting (\( \tau=0.1 \)) to balance minimal creative variation with strong factual consistency.

In our implementation, we utilize the chat template to standardize input-output formats. The system prompt designates the model as a "professional editor specializing in rewriting traffic accident reports," with explicit preservation requirements communicated through detailed guidelines. These include maintaining all factual information (times, dates, locations, vehicle details), removing uninformative placeholders (e.g., "nan," "unknown"), preserving chronological order, and employing consistent professional language. The augmentation process handles batch processing of instances for computational efficiency.

This methodology delivers three significant improvements over conventional preprocessing approaches: standardization of linguistic variability across temporal and jurisdictional dimensions; enhancement of narrative fluidity without introducing factual distortion; and preservation of complex crash dynamics involving human, vehicular, and environmental factors. By maintaining semantic integrity while improving textual coherence, our technique creates crash narratives that are more consistent and amenable to downstream natural language processing tasks.

\subsection{Supervised Fine-tuning of LLM}
During the fine-tuning phase, the traffic crash severity inference task is framed as a next-token generation task. This process can be described as~\cite{shiediting}:

\begin{equation}
p_{\theta}(T_i) = \prod_{j=1}^{|T_i|} p_{\theta}(t_j^{(i)}|t_1^{(i)}, \cdots, t_{j-1}^{(i)}),
\end{equation}

where $T_i$ is the $i$-th example in the training data, $p_\theta$ is the LLM model, $t_j^{(i)}$ denotes the $j$-th token in $T_i$. 

The LLM's parameters are fine-tuned by maximizing the likelihood $p_\theta(T) = \prod_{i=1}^N p_\theta(T_i)$. Both the system prompt and the user prompt are masked for loss computation during training~\cite{shi2025enhancing}. In our setting, the model is fine-tuned to predict crash severity levels, a task that demands a deep understanding of crash dynamics.

\subsection{Gradient-based Explanation}
\subsubsection{Gradient-based Attribution}
\label{section:gradient}
To ensure trustworthiness, we seek to understand the complex inner workings of our supervised fine-tuned LLM in the traffic safety domain by applying a gradient-based explanation method originally developed for analyzing the impact of instruction tuning on language models 
\cite{wu2024language}. 

Gradient-based attribution techniques, which have been widely used to explain deep learning models \cite{sundararajan2017axiomatic}, can help identify which words in an input text have the greatest influence on the model's output. In the context of traffic safety, this could enable pinpointing specific terms or phrases in incident reports that the model deems most indicative of safety issues. Following \cite{wu2024language}, we will use a first-order Taylor approximation of the difference in output probabilities when including or excluding each input token. Normalizing and thresholding these attribution scores yield a robust measure of each input token's importance. The importance $I_{n,m}$ of input token $x_n$ to output token $y_m$ is defined as:

\begin{equation}
I_{n,m} = p(y_m | Z_m) - p(y_m | Z_{m, /n})
\end{equation}
where $Z_m$ is the context for generating $y_m$ consisting of the concatenation of prompt $X$ and the first $m-1$ tokens of response $Y$, and $Z_{m, /n}$ omits token $x_n$ from $Z_m$. This is approximated using a first-order Taylor expansion:

\begin{equation}
I_{n,m} \approx \left\langle \frac{\partial f(y_m|Z_m)}{\partial E_i[x_n]}, E_i[x_n] \right\rangle
\end{equation}
where $E_i[x_n]$ is the input word embedding of token $x_n$ extracted from the fine-tuned LLM. The normalized pairwise importance score $\hat{S}_{n,m}$ is then defined as:
\begin{equation}
    \hat{S}_{n,m} =
        \begin{cases}
            \left\lceil\frac{L \times I_{n,m}}{\max_{n'}^{N} I_{n',m}}\right\rceil & \text{if $\left\lceil\frac{L \times I_{n,m}}{\max_{n'}^{N} I_{n',m}}\right\rceil > b$} \cr
            0 & \text{otherwise}
        \end{cases}
\end{equation}
where a scaling factor $L$ and a binary threshold $b$ are hyperparameters. In this study, we set $L= 100$ and $b =1$.

\subsubsection{Crash Severity Analysis with Word-level Attribution}
\label{section:gptsummary}

The foundation of our analytical framework relies on word-level attribution, which assigns importance scores to individual words or phrases within crash narratives based on the gradient attribution, described previously in Section ~\ref{section:gradient}. The attribution scores quantify each token's contribution to the final prediction. Higher attribution scores suggest stronger associations with crash severity outcomes. 

To enable aspect-aware explanation, our methodology decomposes crash narratives with word-level attribution into five key categories of contributing factors: 1) environmental conditions (weather, lighting, road surface quality, etc.); 2) vehicle and occupant characteristics (vehicle types, protection systems, etc.); 3) driver behavioral elements (speed, intoxication, maneuvers, etc.); 4) infrastructure features (road design, traffic control devices); and 5) unusual aspects with unexpectedly high attribution scores that may represent unique contributing factors. This multi-aspect approach enables comprehensive assessment of crash severity determinants.

The implementation consists of a semi-automated pipeline utilizing GPT-4o \cite{openai2024gpt4o} to process and interpret word-level attribution data. Raw crash narratives, augmented with attribution scores, undergo systematic analysis through a carefully engineered prompt structure. This prompt directs the language model to process text with embedded attribution values, identify high-scoring words relevant to each factor category, and provide concise analytical summaries while maintaining output consistency.

To facilitate systematic analysis, we standardized the output using a JSON structure that preserves organizational consistency across all analyzed reports. For each factor category, the output includes a narrative summary capturing key insights and an array of high-scoring words with their associated attribution values. This structured approach enables qualitative assessment through the summaries as well as quantitative analysis via the attribution scores.

The core advantage of this approach lies in the combination of domain-specific prompting with robust natural language processing capabilities of LLMs, resulting in precisely formatted analyses that highlight each crash's salient risk factors in a transparent, interpretable manner. It advances crash severity analysis by leveraging the linguistic pattern recognition capabilities of LLMs while maintaining analytical rigor through attribution-based evidence.

\section{CrashSage: Crash Severity Inference }
In this study, we focus on fine-tuning a LLM for the task of crash severity inference, aligning a general-purpose LLM with domain-specific knowledge in road safety.

\subsection{Supervised Fine-tuning  and Hyperparameters}
We performed supervised fine-tuning on the Llama3-8B model \cite{llama3modelcard} using parameter-efficient fine-tuning via LoRA \cite{hulora}. The model was using AdamW \cite{loshchilovdecoupled} as optimizer and trained for 30 epochs using the DeepSpeed framework \cite{rasley2020deepspeed}. We employed a LoRA configuration with rank (r) of 128, alpha scaling factor of 256, and dropout rate of 0.1, targeting all linear layers in the model. Training was conducted with a learning rate of 3e-5 using a cosine scheduler with 5\% warmup, weight decay of 1e-4, and maximum gradient norm of 1.0. For optimization efficiency, we utilized gradient checkpointing and accumulated gradients over 16 steps with a per-device batch size of 1. The model processed sequences with a maximum length of 2,048 tokens and was trained in bfloat16 precision. This configuration balances computational efficiency with effective knowledge transfer while maintaining reasonable memory requirements. The experiments are conducted on a server  with four Nvidia A6000 48GB GPUs.

\subsection{Baseline Methods}

This study establishes several baseline configurations to evaluate the performance of our supervised fine-tuned LLaMA3-8B \cite{llama3modelcard} model against state-of-the-art proprietary models for the traffic crash severity classification task. We implemented three distinct prompting strategies with GPT-4o \cite{openai2024gpt4o} as our primary baselines, while also extending our evaluation to include LLaMA3-8B, LLaMA3-70B, and GPT-4o mini \cite{openai2024gpt4omini} for broader comparative analysis.

The baselines are structured to assess how different prompting techniques influence model performance in classifying traffic crash severity into two severity levels: ``No apparent or minor injury'' or ``Serious injury or fatal.'' These baselines provide crucial reference points for evaluating the efficacy of our supervised fine-tuning approach against powerful foundation models using various prompting strategies.

\begin{table}
  \caption{Experimental Settings and Abbreviations\label{tab:experiments}}
  \centering
  \begin{tabular}{lp{12cm}}
  \toprule
  \multicolumn{2}{l}{\textbf{Approaches}} \\
  \midrule
  Zero-shot& Models evaluated without any examples \\
  Zero-shot with Chain-of-Thought& Models prompted to explain reasoning step-by-step \\
  Few-shot& Models provided with a small number of examples \\
  Supervised Fine-Tuning& Model fine-tuned on task-specific data \\
  \midrule
  \multicolumn{2}{l}{\textbf{Models Evaluated}} \\
  \midrule
  LLaMA3-8B \cite{llama3modelcard} & 8 billion parameter open-source model \\
  LLaMA3-70B \cite{llama3modelcard}& 70 billion parameter open-source model \\
  GPT-4o-mini \cite{openai2024gpt4omini} & gpt-4o-mini-2024-07-18, multimodal model from OpenAI \\
  GPT-4o \cite{openai2024gpt4o} & gpt-4o-2024-11-20, advanced multimodal model from OpenAI \\
  \midrule
  \multicolumn{2}{l}{\textbf{Sampling Strategy}} \\
  \midrule
  Sampling Strategy & Greedy decoding for all models \\
  \bottomrule
  \end{tabular}
  \end{table}

\subsubsection{Zero-Shot Prompting}

Our first baseline employs a zero-shot prompting strategy, wherein models receive a concise instruction without examples. The model is prompted with domain-specific context identifying it as a professional road safety engineer and tasked with classifying crash severity based solely on the provided crash description. The prompt used is shown below:

\begin{quote}
\textit{You are a professional road safety engineer.}

\textit{You are given a detailed description for a traffic crash.}

\textit{Please classify the severity of the crash into one of two categories: 'No apparent or minor injury', 'Serious injury or fatal accident'.}

\textit{You can only output one of the classification result in your answer.}
\end{quote}

This approach evaluates the model's inherent ability to perform the classification task without prior examples, relying exclusively on its pre-trained knowledge on traffic safety.

\subsubsection{Zero-Shot Chain-of-Thought}

The second baseline implements a zero-shot chain-of-thought (CoT) approach, which extends the zero-shot prompting by explicitly instructing the model to analyze the traffic crash before outputting the classification result. The prompt used for this approach is:

\begin{quote}
\textit{You are a professional road safety engineer.}

\textit{You are given a detailed description for a traffic crash.}

\textit{Please analyze this traffic crash with careful reasoning first, and then classify the severity of the crash into one of the two categories: 'No apparent or minor injury', 'Serious injury or fatal accident'.}

\textit{You can only output one of the classification result at the end of your answer.}
\end{quote}

This modification encourages the model to engage in a more deliberate reasoning process, potentially leading to improved decision-making. The CoT approach is designed to assess whether explicit instructions for analytical reasoning enhance classification accuracy compared to direct zero-shot prompting.

\subsubsection{Few-Shot Learning}

The third baseline utilizes a few-shot learning paradigm, presenting the model with two exemplar traffic crashes, one in each severity category, and their corresponding severity outcomes prior to requesting classification of the target case. The prompt includes carefully selected examples that serve as implicit demonstrations of the reasoning process and decision criteria:

\begin{quote}
\textit{You are a professional road safety engineer.\\}

\textit{Here are two examples of traffic crashes and their severity classification:}

\textit{[Example 1 with label 'No apparent or minor injury']}

\textit{No apparent or minor injury}

\textit{[Example 2 with label 'Serious injury or fatal']}

\textit{Serious injury or fatal\\}

\textit{You are given a detailed description for a traffic crash.}

\textit{Please classify the severity of the crash into one of two categories: 'No apparent or minor injury', 'Serious injury or fatal'.}

\textit{You can only output one of the classification result in your answer.}
\end{quote}

The few-shot approach leverages LLMs' in-context learning capabilities to learn from a minimal set of examples and apply that knowledge to new cases, a strategy  demonstrated to improve performance across a range of natural language processing tasks.

\subsection{Results}

\begin{table}[h]
    \centering
    \caption{Performance comparison of different models across evaluation metrics.}
    \label{tab:model_performance}
    \begin{tabular}{llcccc}
        \toprule
        Setting & Model & Macro-F1 & Accuracy & Macro-Recall & Macro-Precision \\
        \midrule
        \multirow{4}{*}{ZS} 
        & LLaMA3-8B & 0.6726 & 0.6883 & 0.67 & 0.68 \\
        & LLaMA3-70B & 0.6345 & 0.6355 & 0.67 & 0.67 \\
        & GPT-4o-mini & 0.6711 & 0.7078 & 0.67 & 0.72 \\
        & GPT-4o & 0.7067 & 0.7229 & 0.70 & 0.72 \\
        \midrule
        \multirow{4}{*}{ZS\_CoT} 
        & LLaMA3-8B & 0.5071 & 0.5346 & 0.59 & 0.66 \\
        & LLaMA3-70B & 0.6059 & 0.6099 & 0.65 & 0.67 \\
        & GPT-4o mini & 0.3717 & 0.4413 & 0.52 & 0.55 \\
        & GPT-4o & 0.3693 & 0.4443 & 0.52 & 0.58 \\
        \midrule
        \multirow{4}{*}{FS} 
        & LLaMA3-8B & 0.6851 & 0.6867 & 0.70 & 0.69 \\
        & LLaMA3-70B & 0.7051 & 0.7184 & 0.70 & 0.71 \\
        & GPT-4o-mini & 0.6560 & 0.6898 & 0.66 & 0.69 \\
        & GPT-4o & 0.7062 & 0.7259 & 0.70 & 0.72 \\
        \midrule
        SFT & LLaMA3-8B & \textbf{0.7361} & \textbf{0.7395} & \textbf{0.74} & \textbf{0.74} \\
        \bottomrule
    \end{tabular}
\end{table}

In evaluating the outcomes of our experimental comparisons, it is evident that the approach involving supervised fine-tuning (SFT) of LLaMA3-8B consistently provides superior performance across multiple metrics when compared to the zero-shot (ZS), zero-shot and chain-of-thought (ZS\_CoT), and few-shot (FS) settings for unadapted models. As summarized in Table \ref{tab:model_performance}, SFT demonstrates a Macro-F1 score of 0.7361, surpassing all baseline methods and highlighting the benefits of specialized domain adaptation. These findings derive in part from the motivations that guided our experiment design, namely the hypothesis that traffic crash narratives require careful domain grounding and that purely general-purpose prompting may not suffice in capturing nuanced, context-sensitive crash conditions.

A closer look at the ZS and FS settings indicates subtle differences. GPT-4o shows stronger performance in the zero-shot condition than the other baseline models, most notably in Macro-F1 (0.7067) and accuracy (0.7229). However, when LLaMA3-8B is given a minimal number of examples in the few-shot setting, its performance moves closer to that of GPT-4o, demonstrating the model's capacity to incorporate contextual cues with only limited domain examples. The results in the ZS\_CoT setting indicate that chain-of-thought prompting does not uniformly enhance outcomes; the performance declines for some models. This pattern suggests that while reflective reasoning can be beneficial in some contexts, introducing additional reasoning steps here may pose risks of over-interpreting or extraneous text generation that does not align with domain-consistent interpretations of crash data.

A point of particular interest is the comparable or even superior performance of fine-tuned LLaMA3-8B compared to a significantly larger model like LLaMA3-70B. The smaller model, when tuned on domain-specific data, not only reduces computational overhead but also captures key details of traffic crash environments, including interactions among vehicles, environmental features, and driver attributes. This highlights the potential of fine-tuned smaller models to perform competitively in specialized domains, offering an efficient alternative to large-scale models.

These results validate our initial hypothesis that transforming structured data into domain-focused narratives prior to model tuning enhances performance. The observed gain from SFT supports the notion that foundation models benefit from specialized retraining to internalize the nuanced complexity of traffic crashes. Moreover, the varying performance across prompting strategies underscores the importance of aligning LLM behavior with domain-specific reasoning paradigms. Through a systematic comparison of zero-shot, chain-of-thought, few-shot, and SFT approaches, our experiments demonstrate that, while general-purpose LLMs show strong baseline competence, supervised fine-tuning more effectively captures crash-specific scenarios. This finding charts a clear path for future work, where incorporating additional tuning data or exploring even more refined prompting techniques could further enhance both predictive accuracy and explainability in real-world traffic safety applications.

\section{CrashSage: Interpretability and Attribution}
\label{sec:interpretability}
In traffic safety modeling and analysis, interpretability is crucial for building trust and understanding the factors driving model predictions. Our CrashSage framework employs gradient-based attribution techniques to highlight the most influential terms in crash narratives that contribute to traffic crash severity inferences. This approach not only enhances transparency but also provides valuable insights into the complex interplay among factors affecting crash outcomes. 

\subsection{Individual Incident Inspection: Word-Level Explanations of an Accident}

Our word-level attribution method aggregates token-level importance scores derived from gradient-based techniques to identify which words and phrases most significantly influence the model's severity classification. In detail, our implementation identifies token boundaries within the LLM's tokenization scheme and combines scores of sub-word tokens into coherent words, providing more intuitive interpretations for domain experts. These importance scores quantify how much each word contributes to the final prediction, enabling analysts to understand which aspects of a crash narrative were most decisive in the model's reasoning process.

\subsubsection{Example 1: No apparent or minor injury crash}

For illustration, consider a \textit{no apparent or minor injury} crash example analyzed by our CrashSage. Figure~\ref{fig:wordlevelattribution} displays the crash narrative text where each word is associated with attribution scores. These scores represent the word-level attribution values derived from our model's analysis. Words receiving higher attribution scores are identified as having greater influence on the model's classification decision regarding crash severity. This word-level attribution approach provides a more interpretable view of which textual elements most significantly contributed to the severity assessment.

\begin{figure}[htbp]
  \centering
  \fbox{%
      \begin{minipage}{0.95\linewidth}
      \vspace{0.5em}
      \small
      \textbf{\textit{Example Narrative with Word-Level Attribution:}} \\
      \textbf{On}\textcolor{gray}{[1.92]} \textbf{June}\textcolor{gray}{[1.96]} \textbf{29,}\textcolor{gray}{[2.66]} \textbf{2022,}\textcolor{gray}{[4.28]} at\textcolor{gray}{[1.68]} 8:00\textcolor{gray}{[1.00]} \textbf{PM,}\textcolor{gray}{[2.00]} a\textcolor{gray}{[1.63]} traffic\textcolor{gray}{[1.00]} accident\textcolor{gray}{[1.68]} occurred\textcolor{gray}{[1.39]} on\textcolor{gray}{[1.00]} Alternate\textcolor{gray}{[1.00]} Route\textcolor{gray}{[1.00]} 097ARi\textcolor{gray}{[1.68]} in\textcolor{gray}{[1.00]} \textbf{Chelan,}\textcolor{gray}{[3.39]} \textbf{Washington.}\textcolor{gray}{[2.48]} 
      
      [...] under \textbf{dusk}\textcolor{gray}{[2.28]} conditions.\textcolor{gray}{[2.00]} [...] rural\textcolor{gray}{[1.00]} \textbf{two-lane}\textcolor{gray}{[2.00]} road\textcolor{gray}{[1.00]} [...] 
      
      [...] an \textbf{[BRAND 1]}\textcolor{gray}{[2.00]} \textbf{[MODEL 1] vehicle}\textcolor{gray}{[3.20]} manufactured\textcolor{gray}{[1.46]} in \textbf{2005,}\textcolor{gray}{[3.68]} [...] \textbf{35-year-old}\textcolor{gray}{[2.68]} male,\textcolor{gray}{[1.68]} [...] time\textcolor{gray}{[0.00]} of\textcolor{gray}{[1.42]} the\textcolor{gray}{[0.00]} \textbf{crash.}\textcolor{gray}{[3.39]} [...]
      
      [...] \textbf{36-year-old}\textcolor{gray}{[3.68]} female,\textcolor{gray}{[0.00]} [...]\textcolor{gray}{[0.00]} of\textcolor{gray}{[0.00]} \textbf{47.9512}\textcolor{gray}{[2.00]} [...] \textbf{hit-and-run}\textcolor{gray}{[3.68]} \textbf{incident.}\textcolor{gray}{[3.11]}
      \vspace{0.5em}
      \end{minipage} }
      \caption{Word-level attribution visualization for a *no apparent or minor injury* crash example. Terms with higher attribution scores (shown in brackets) have greater influence on the model's prediction. High-influence terms are highlighted in \textbf{bold} with color intensity representing attribution strength: \textcolor{gray}{gray} (minimal).}
  \label{fig:wordlevelattribution}
\end{figure}

To better visualize the relative attribution of each word or phrase, we present the results using a color-coded heatmap. To minimize visual clutter, the visualization uses a binary color mapping scheme that clearly differentiate  between high and low levels of attribution significance. Elements with higher attribution values are displayed in red, while those with lower but still notable attribution values appear in green. This divergent color scheme creates an intuitive visual hierarchy that emphasizes textual segments according to their influence on model predictions.

\begin{figure}[htbp]
  \centering
  \includegraphics[width=\textwidth]{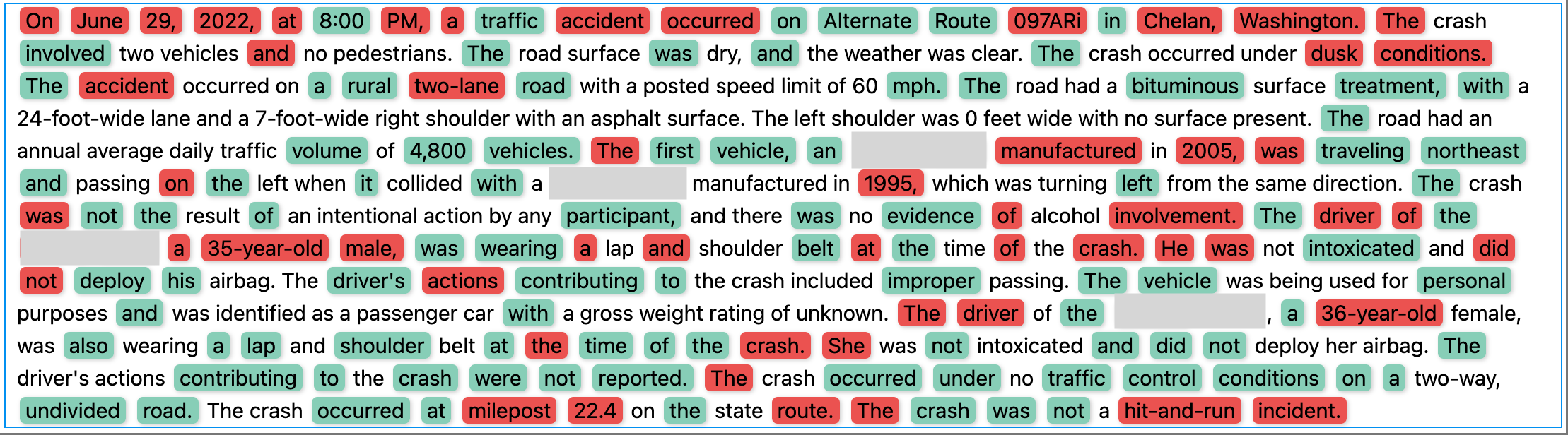}
  \caption{Exemplar visualization of attribution scores for a crash classified as ``No apparent or minor injury.'' Colors indicate attribution strength: red indicates strongly influential words (high positive attribution), while green denotes moderately influential terms. This heat map visualization reveals which textual elements most significantly impact the model's severity classification decision. In this figure, sensitive vehicle information (e.g., make and model) has been redacted using gray boxes. This redaction was performed manually and does not alter or affect the substantive content or analytical integrity of the narrative.}
  \label{fig:attribution_example}
\end{figure}

The attribution visualization in Figure~\ref{fig:attribution_example} reveals several key patterns in how the CrashSage evaluates crash narratives. Temporal markers (``On,'' ``June,'' ``29,'' ``2022,'' ``PM'') receive moderate to high attribution scores, indicating the importance of time-related information in severity assessment. Location identifiers (``Chelan,'' ``Washington'') similarly show strong influence, suggesting geographical context impacts prediction.

Environmental conditions, particularly ``dusk'' and ``two-lane'' road configuration, demonstrate notable attribution weights, aligning with safety research that identifies limited visibility and road type as significant risk factors. Vehicle-specific details (``[BRAND 1] [MODEL 1] vehicle,'' ``2005'') receive high attribution scores, likely reflecting the model's attention to vehicle age and model. The most substantial attributions appear for driver characteristics (``35-year-old male,'' ``36-year-old female'') and incident type indicators, particularly the negated ``hit-and-run incident'' phrase, which appears critical to the model's classification decision.

\definecolor{sectionbg}{RGB}{240,240,240}
\definecolor{factorbg}{RGB}{250,250,250}

We further employ a semi-automated pipeline with GPT-4o, described in section \ref{section:gptsummary},to summarize crash narratives with word-level attribution scores, wherein the system prompt for the LLM specifies a concise JSON-based output structure, organizing into five key categories: environmental, vehicle/occupant, behavioral, infrastructure, and unusual/standout factors. Raw crash text, augmented with word-level attribution scores, is processed by the GPT-4o. The core advantage lies in pairing domain-specific prompting with robust file handling, leading to precisely formatted results that highlight each crash's salient risk factors by categories. A structured summary of the same example is presented below.


\factorsection{Environmental Factors}{The crash occurred on June 29, 2022, at 8:00 PM under clear weather and dry road conditions. It was dusk at the time, and the location was a rural two-lane road in Chelan, Washington.}{
2022 & (4.28) \\
Chelan & (3.39) \\
29 & (2.66) \\
Washington & (2.48) \\
dusk & (2.28) \\
8:00 PM & (2.00) \\
June & (1.96) \\
}

\factorsection{Vehicle and Occupant Factors}{The crash involved two vehicles: a 2005 [BRAND 1] [MODEL 1] driven by a 35-year-old male and a 1995 [BRAND 2] [MODEL 2] driven by a 36-year-old female. Both drivers were wearing seat belts, and no airbags were deployed. Neither driver was intoxicated.}{
2005 & (3.68) \\
36-year-old & (3.68) \\
{[BRAND 1] [MODEL 1] vehicle} & (3.20) \\
1995 & (3.00) \\
35-year-old & (2.68) \\
{[BRAND 2] [MODEL 2] vehicle} & (1.00) \\
}

\factorsection{Driver Behavioral Factors}{The driver of the [BRAND 1] [MODEL 1] vehicle was improperly passing when the crash occurred. There was no evidence of alcohol involvement, and the crash was not intentional.}{
no evidence of alcohol involvement & (2.00) \\
improper passing & (1.00) \\
not intentional & (1.00) \\
}

\factorsection{Infrastructure Factors}{The crash occurred on Alternate Route 097ARi, a rural two-lane road with a posted speed limit of 60 mph. The road had a bituminous surface treatment, a 24-foot-wide lane, and a 7-foot-wide right shoulder. The left shoulder was 0 feet wide. The location had an AADT of 4,800 vehicles and no traffic control.}{
two-lane & (2.00) \\
Alternate Route 097ARi & (1.68) \\
AADT 4,800 & (1.00) \\
no traffic control & (1.00) \\
}

\factorsection{Unusual/Standout Factors}{The crash was not a hit-and-run incident, and the specific location at milepost 22.4.}{
hit-and-run & (3.68) \\
milepost 22.4 & (2.00) \\
}
  
\
\subsubsection{Example 2: Serious injury and fatal crash}
\begin{figure}[htbp]
  \centering
  \fbox{%
      \begin{minipage}{0.95\linewidth}
      \vspace{0.5em}
      \small
      \textbf{\textit{Example Narrative with Word-Level Attribution (Serious/Fatal Crash):}} \\
      On\textcolor{gray}{[0.00]} September\textcolor{gray}{[1.22]} \textbf{5,}\textcolor{gray}{[2.68]} \textbf{2020,}\textcolor{gray}{[3.61]} at\textcolor{gray}{[1.68]} \textbf{1:00}\textcolor{gray}{[2.62]} \textbf{PM,}\textcolor{gray}{[3.02]} a\textcolor{gray}{[1.00]} \textbf{rear-end}\textcolor{gray}{[2.28]} collision\textcolor{gray}{[1.92]} occurred\textcolor{gray}{[1.00]} on\textcolor{gray}{[1.63]} State\textcolor{gray}{[0.00]} Route\textcolor{gray}{[1.00]} \textbf{542i}\textcolor{gray}{[3.11]} \textbf{(MAINLINE)}\textcolor{gray}{[4.37]} in\textcolor{gray}{[0.00]} \textbf{Whatcom,}\textcolor{gray}{[3.00]} Washington.\textcolor{gray}{[2.00]} 
      
      [...] \textbf{two-lane}\textcolor{gray}{[2.00]} road\textcolor{gray}{[1.68]} [...] \textbf{2001}\textcolor{gray}{[2.00]} [BRAND 1] \textcolor{gray}{[0.00]} \textbf{[MODEL 1] vehicle,}\textcolor{gray}{[3.05]} [...] \textbf{BRAND 2}\textcolor{gray}{[0.00]} \textbf{MODEL 2}\textcolor{gray}{[3.00]} motorcycle,\textcolor{gray}{[2.00]} [...]
      
      [...] \textbf{motorcycle's}\textcolor{gray}{[2.68]} rear-end\textcolor{gray}{[0.00]} collision\textcolor{gray}{[1.00]} [...] alcohol\textcolor{gray}{[0.00]} involvement.\textcolor{gray}{[2.00]} [...] [BRAND 1] \textcolor{gray}{[0.00]} \textbf{MODEL 2}\textcolor{gray}{[4.31]} [...] \textbf{intoxication}\textcolor{gray}{[2.68]} [...] actions.\textcolor{gray}{[2.00]} [...]
      
      [...] \textbf{milepost}\textcolor{gray}{[3.36]} \textbf{35.42.}\textcolor{gray}{[4.52]}
      \vspace{0.5em}
      \end{minipage} }
      \caption{Word-level attribution visualization for a \textit{serious injury or fatal} crash example. Terms with higher attribution scores (shown in brackets) have greater influence on the model's prediction. High-influence terms are highlighted in \textbf{bold} with attribution strength represented by numerical values. This visualization demonstrates how the model attends to different factors when assessing a severe crash compared to minor injury cases.}
  \label{fig:serious_attribution}
\end{figure}

\begin{figure}[htbp]
  \centering
  \includegraphics[width=\textwidth]{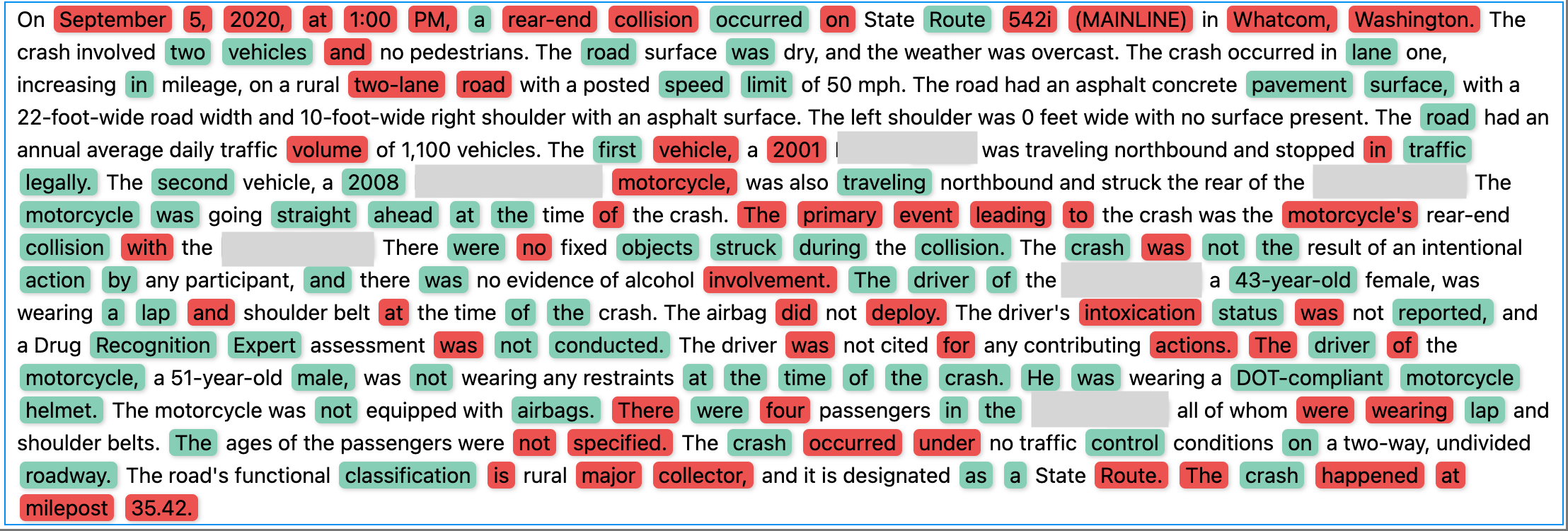}
  \caption{Example visualization of attribution scores for a crash classified as ``Serious injury or fatal accident.'' Colors indicate attribution strength: red indicates strongly influential words (high positive attribution), while green denotes moderately influential terms. Notable high-attribution elements include specific temporal and location identifiers (``2020,'' ``PM,'' ``542i,'' ``MAINLINE''), vehicle details (``[MODEL 1],'' ``motorcycle''), and precise geographic locations. In this figure, sensitive vehicle information (e.g., make and model) has been redacted using gray boxes. This redaction was performed manually and does not alter or affect the substantive content or analytical integrity of the narrative.}
  \label{fig:serious_attribution_heatmap}
\end{figure}

Similarly, the attribution visualization in Figure~\ref{fig:serious_attribution_heatmap} reveals distinctive patterns for this serious/fatal crash. While temporal and location information remain important as in the minor injury case, this example shows particularly strong attribution to specific roadway identifiers (``542i,'' ``MAINLINE'') and precise geographic coordinates. The motorcycle involvement (``[MODEL 2],'' ``motorcycle's'') receives substantial weight, consistent with traffic safety research identifying motorcycles as associated with higher crash severity outcomes. Notable also is the high attribution for ``intoxication'' even though it's negated in the text, suggesting the model considers this factor critically important when assessing crash severity.

The attribution pattern in this serious/fatal crash example reveals notable differences compared to the minor injury case previously analyzed. While both examples show significant attribution to temporal and location information, this severe crash narrative demonstrates particularly strong influence from specific roadway identifiers (``542i,'' ``MAINLINE'') and precise geographic ``milepost 35.42''. 

The motorcycle involvement receives substantial attribution weight, with high scores for ``[MODEL 1][3.00]'' and ``motorcycle's[2.68],'' aligning with traffic safety research that consistently identifies motorcycles as associated with increased crash severity. Vehicle type information (``[BRAND 1] [MODEL 1][4.31]'') shows remarkably high attribution, possibly indicating the model has learned relationships between vehicle types and crash outcomes from its training data.

Noteworthy is the high attribution score for ``intoxication[2.68]'' despite it being negated in the text, suggesting the model places significant importance on this factor's presence or absence when assessing crash severity. The phrase ``rear-end[2.28]'' collision type also receives substantial attribution, reflecting its relevance to severity outcomes in motorcycle-involved crashes.

This gradient-based attribution analysis demonstrates how the CrashSage identifies distinctive risk patterns when evaluating crash severity. For serious/fatal crashes, the model focuses intensely on vehicle types (particularly motorcycles), specific road identifiers, and precise location data, while utilizing different attribution patterns for minor injury incidents. The visualization provides transparent insight into the model's decision-making process, revealing how it integrates multiple contextual elements: temporal, environmental, vehicular, and human factors, rather than focusing on isolated aspects. This comprehensive approach mirrors the multifaceted evaluation process used by human safety experts, while offering computational precision in identifying combinations of factors that contribute to different crash outcomes.

Similarly, the semi-automated pipeline is applied to summarize this serious injury/fatal crash narrative, with the resulting output shown below. 

\factorsection{Environmental Factors}{The crash occurred on a dry road surface under overcast weather conditions during the daytime (1:00 PM) on a rural two-lane road in Whatcom, Washington.}{
2020 & (3.61) \\
PM & (3.02) \\
Whatcom & (3.00) \\
5 & (2.68) \\
1:00 & (2.62) \\
Washington & (2.00) \\
September & (1.22) \\
}

\factorsection{Vehicle and Occupant Factors}{The crash involved a 2001 [BRAND 1] [MODEL 1] and a 2008 [BRAND 2] [MODEL 2] motorcycle. The [BRAND 1] [MODEL 1] vehicle had a 43-year-old female driver and four passengers, all wearing seat belts. The motorcycle was driven by a 51-year-old male wearing a DOT-compliant helmet but no other restraints. The [BRAND 1] [MODEL 1]'s airbag did not deploy.}{
{[BRAND 1]} & (3.05) \\
{[BRAND 2]} & (3.00) \\
{[MODEL 1]} & (3.00) \\
intoxication & (2.68) \\
{[MODEL 2]} & (2.36) \\
2001 & (2.00) \\
airbag & (1.65) \\
2008 & (1.00) \\
}

\factorsection{Driver Behavioral Factors}{The motorcycle rear-ended the [BRAND 1] [MODEL 1], which was legally stopped in traffic. There was no evidence of alcohol involvement or intentional actions by either driver. The motorcycle driver was not cited for any contributing actions.}{
motorcycle's & (2.68) \\
intoxication & (2.68) \\
rear-end & (2.28) \\
contributing actions & (2.00) \\
collision & (1.92) \\
}

\factorsection{Infrastructure Factors}{The crash occurred on State Route 542i, a rural two-lane road with a posted speed limit of 50 mph. The road had an asphalt concrete surface and an annual average daily traffic volume of 1,100 vehicles.}{
MAINLINE & (4.37) \\
State Route 542i & (3.11) \\
State Route & (2.59) \\
two-lane & (2.00) \\
volume & (1.68) \\
}

\factorsection{Unusual/Standout Factors}{The crash location's milepost were highly detailed and had high attribution scores.}{
35.42 & (4.52) \\
milepost & (3.36) \\
}

\subsection{Co-occurrence analysis of high-score factors from different aspects}

To reveal relationships between influential crash factors, we conducted analysis,  focusing on high-attribution elements identified through our gradient-based attribution approach. The process begins with factor extraction,  constrained to the top five factors for each of the four aspects: environmental, driver behavioral, vehicle/occupant, and infrastructure.  Subsequently, co-occurring factor pairs are identified and  visualized using a Sankey diagram.  

Semantic grouping was applied to consolidate conceptually similar terms that appeared with varied phrasing throughout the dataset. For instance, different temporal references were unified under broader descriptors such as "time of day". This semantic grouping process enhanced interpretability while preserving the underlying semantic relationships between factors across various crash instances.

The resulting data structure captures the co-occurrence patterns in reference to four fundamental safety aspects (i.e., environmental conditions, driver behaviors, vehicle attributes, and infrastructure features), with each node representing a distinct factor and connecting links quantifying co-occurrence frequency between factor pairs. This graphic visualization reveals how these specific aspects interact in complex crash scenarios. The structure illuminates critical interdependencies, such as correlations between alcohol-related behaviors and specific temporal or roadway characteristics, providing a holistic view of crash dynamics. 

\begin{figure}[htbp]
  \centering
  \includegraphics[width=\textwidth]{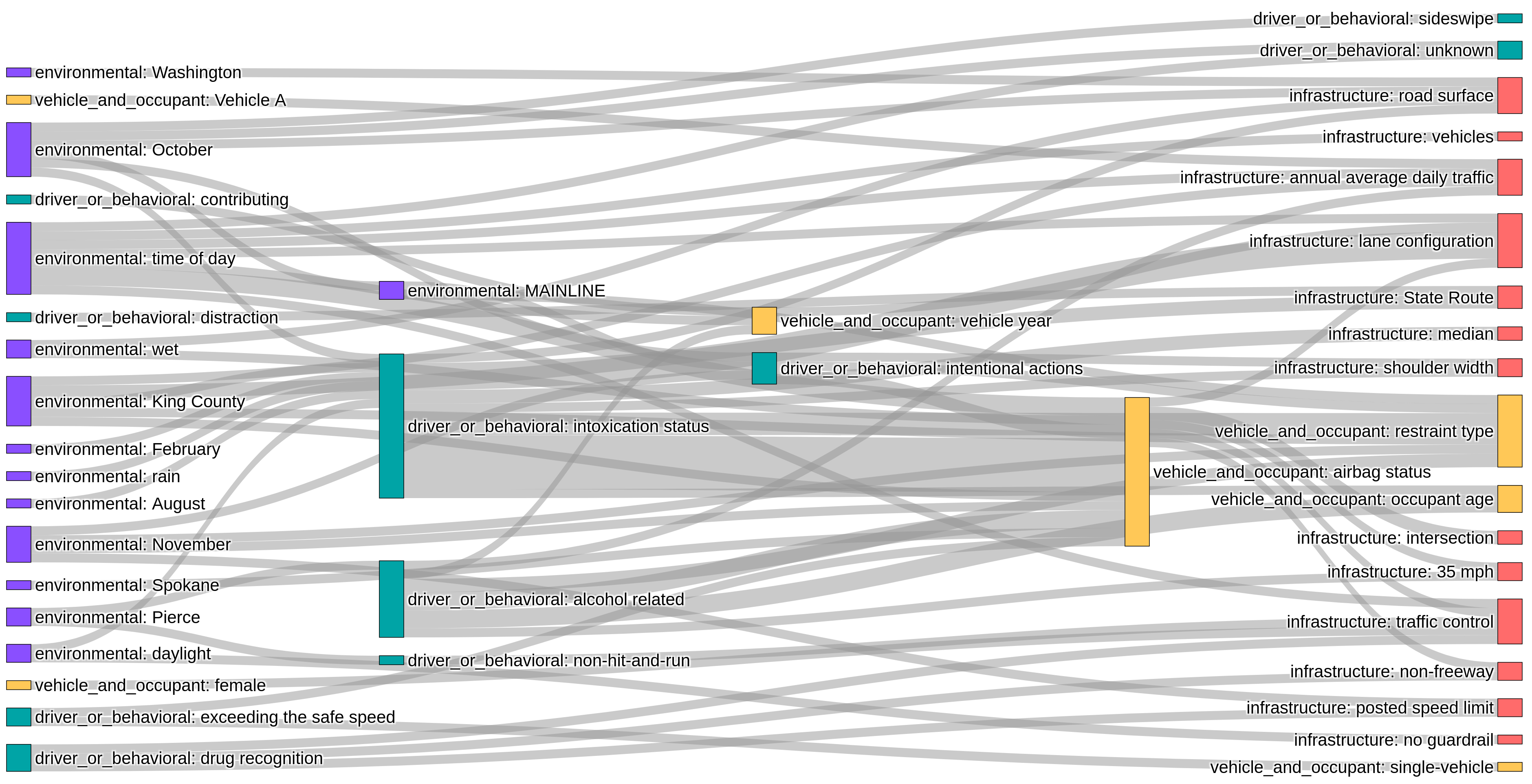}
  \caption{Aspect-level risk factor association visualized via a Sankey diagram. This visualization shows the relationships between identified high-attribution terms and their categorization into broader safety aspects. The thickness of each link indicates the relative frequencies of co-occurrence  between factor categories.}
  \label{fig:sankey_risk_factors}
\end{figure}

Figure \ref{fig:sankey_risk_factors} presents these relationships through a Sankey diagram where safety aspects are color-coded for clarity, and connection strengths are represented by flow widths proportional to co-occurrence frequency. The diagram reveals complex interdependencies among crash factors spanning environmental, behavioral, vehicle/occupant, and infrastructure aspects. Notably, driver behavioral factors, particularly intoxication status and alcohol-related impairment, emerge as central nodes with extensive links to both environmental and infrastructure elements, underscoring the critical role of driver condition in crash severity. Temporal factors, such as time of day, show strong associations with impairment-related behaviors, indicating diurnal patterns in high-risk driving. Additionally, geographic variability is reflected in distinct connection patterns between behavioral factors and specific locations, pointing to regional heterogeneity in crash factor distributions.

Several factor combinations stand out as particularly relevant to crash severity outcomes. The co-occurrence pathway between alcohol-related impairment and excessive speed suggests a synergistic relationship that likely amplifies crash energy and impact forces. Links between intoxication status and restraint utilization indicate a potential behavioral coupling that compounds injury risk through both increased crash likelihood and decreased protection. The visualization also reveals how certain environmental conditions interact with infrastructure characteristics, particularly evident in connections between wet weather and specific road surface attributes, which together create compounded effects on vehicle handling and control.

Vehicle and occupant factors primarily function as intermediate outcome nodes within the graph. Restraint systems and airbag status appear as downstream factors from driver behaviors, suggesting that safety equipment utilization is influenced by driver characteristics. The connection between vehicle year and restraint technologies indicates the progressive integration of advanced occupant protection systems. The position of restraint-related factors as major nodes with multiple connections across the graph underscores their persistent influence across diverse crash scenarios, confirming their fundamental role in severity mitigation regardless of initiating factors.

Infrastructure elements exhibit multifaceted interactions with both antecedent and outcome variables. Lane configuration, with its numerous cross-category connections, exemplifies the complex role of road design in shaping crash dynamics. Road surface characteristics serve as bridging factors between environmental conditions and driver behaviors, suggesting a mediating influence on crash risk. Additionally, traffic control systems display strong association with driver decision-making, reflecting their impact on behavioral responses. Collectively, these interconnections highlight how the roadway environment influences and responds to driver behavior patterns, creating feedback loops that affect crash outcomes.

This diagram analysis demonstrates that crash severity outcome arise from intricate interactions among  diverse factors across different aspects. The co-occurrence patterns suggest that effective traffic safety interventions require multifaceted approaches addressing both human factors and infrastructure design. Special attention is warranted for central behavioral nodes, which act as critical links between environmental conditions and crash outcomes.

Rather than viewing risk factors in isolation, the sankey diagram highlights their interrelationships and potential synergistic effects on crash severity outcomes. This integrated perspective offer road safety researchers and practitioners with an evidence-based foundation for developing multifaceted interventions that account for interconnected dynamics of crash development, thereby supporting more effective and systematic improvements in transportation systems.

\section{Conclusion and Discussions}
This paper introduces CrashSage, a novel LLM-based framework designed to address key challenges in traffic safety analysis. By transforming traditional crash records into coherent textual narratives, followed by context-aware data augmentation, our approach significantly mitigates information loss inherently associated with conventional tabular representations. Building upon enriched narratives, we fine-tune the LLaMA3-8B model to infer crash severity while generating interpretable insights grounded in domain-specific context.

Our experimental results show that the supervised fine-tuned LLaMA3-8B model outperforms comparative baselines, including zero-shot, zero-shot chain-of-thought, and few-shot prompting strategies. Moreover, the integration of gradient-based attribution methods enhances interpretability by uncovering the complex interplay among crash-related factors. This enables domain experts and decision-makers to identify the most influential elements (e.g., speed, intoxication, infrastructure). This level of interpretability is especially valuable for informing targeted traffic safety interventions.

Despite its demonstrated strengths, we acknowledge several limitation of the proposed framework. As a purely text-based approach, it cannot capture other modalities, such as visual cues from roadside cameras or vehicle dash-cam footage, which can be crucial for understanding the time-critical driver action or vehicle status.  Additionally, while our gradient-based explanations offer transparency, they rely on approximations of model behavior and may not fully capture the full depth of the model's internal representations or latent reasoning processes.

Looking ahead, we identify three promising directions for future work. First, integrating video and sensor data into the LLM pipeline could enhance crash narratives with real-time spatiotemporal context, improving model accuracy and robustness. Second, incorporating interpretability-driven constraints into the fine-tuning process may strengthen the consistency and reliability of explanations across diverse scenarios. Third, extending CrashSage from retrospective analyses to proactive risk estimation, where evolving traffic events trigger real-time alerts, offers significant operational benefits if deployed across large-scale transportation networks.

In summary, our work demonstrates the potential of LLMs to bridge structured data with natural language reasoning for  advancing traffic safety research. By improving modeling accuracy, interpretability, and real-time applicability, CrashSage lays the groundwork for a more insightful, transparent, and actionable approach to crash analysis, empowering transportation agencies to make informed, data-driven decisions aimed at reducing road injuries and fatalities.

\bibliographystyle{unsrt}  
\bibliography{references}  
\end{document}